\documentclass[conference]{IEEEtran}
\IEEEoverridecommandlockouts
\usepackage{cite}
\usepackage{amsmath,amssymb,amsfonts}
\usepackage{algorithmic}
\usepackage{graphicx}
\usepackage{textcomp}
\usepackage{xcolor}
\usepackage{makecell}
\usepackage{listings}

\colorlet{punct}{red!60!black}
\definecolor{background}{HTML}{EEEEEE}
\definecolor{delim}{RGB}{20,105,176}
\colorlet{numb}{magenta!60!black}

\lstdefinelanguage{json}{
    basicstyle=\tiny\ttfamily, 
    numbers=left,
    numberstyle=\scriptsize,
    stepnumber=1,
    numbersep=8pt,
    showstringspaces=false,
    breaklines=true,
    frame=lines,
    xleftmargin=4em,
    framexleftmargin=4em,
    xrightmargin=-4em,
    framexrightmargin=-4em,
    literate=
     *{0}{{{\color{numb}0}}}{1}
      {1}{{{\color{numb}1}}}{1}
      {2}{{{\color{numb}2}}}{1}
      {3}{{{\color{numb}3}}}{1}
      {4}{{{\color{numb}4}}}{1}
      {5}{{{\color{numb}5}}}{1}
      {6}{{{\color{numb}6}}}{1}
      {7}{{{\color{numb}7}}}{1}
      {8}{{{\color{numb}8}}}{1}
      {9}{{{\color{numb}9}}}{1}
      {:}{{{\color{punct}{:}}}}{1}
      {,}{{{\color{punct}{,}}}}{1}
      {\{}{{{\color{delim}{\{}}}}{1}
      {\}}{{{\color{delim}{\}}}}}{1}
      {[}{{{\color{delim}{[}}}}{1}
      {]}{{{\color{delim}{]}}}}{1},
}

\newcommand{\algname}{PAIRS AutoGeo }
\newcommand{\algnamens}{PAIRS AutoGeo}

\def\BibTeX{{\rm B\kern-.05em{\sc i\kern-.025em b}\kern-.08em
    T\kern-.1667em\lower.7ex\hbox{E}\kern-.125emX}}
\begin{document}

\title{\algnamens: an Automated Machine Learning Framework for Massive Geospatial Data
}

\author{\IEEEauthorblockN{Wang Zhou}
\IEEEauthorblockA{\textit{IBM T.J. Watson Research Center} \\
\textit{IBM Research}\\
Yorktown Heights, New York\\
wang.zhou@ibm.com}
\and
\IEEEauthorblockN{Levente J. Klein}
\IEEEauthorblockA{\textit{IBM T.J. Watson Research Center} \\
\textit{IBM Research}\\
Yorktown Heights, New York\\
kleinl@us.ibm.com}
\and
\IEEEauthorblockN{Siyuan Lu}
\IEEEauthorblockA{\textit{IBM T.J. Watson Research Center} \\
\textit{IBM Research}\\
Yorktown Heights, New York\\
lus@us.ibm.com}
}

\maketitle
\IEEEpubidadjcol

\begin{abstract}
An automated machine learning framework for geospatial data named \emph{\algnamens} is introduced on IBM PAIRS Geoscope big data and analytics platform. The framework simplifies the development of industrial machine learning solutions leveraging geospatial data to the extent that the user inputs are minimized to merely a text file containing labeled GPS coordinates. \algname automatically gathers required data at the location coordinates, assembles the training data, performs quality check, and trains multiple machine learning models for subsequent deployment. The framework is validated using a realistic industrial use case of tree species classification. Open-source tree species data are used as the input to train a random forest classifier and a modified ResNet model for 10-way tree species classification based on aerial imagery, which leads to an accuracy of 59.8\% and 81.4\%, respectively. This use case exemplifies how \algname enables users to leverage machine learning without extensive geospatial expertise.


\end{abstract}


\section{Introduction}


Continuous monitoring of arbitrary location on Earth has become a reality with increasing availability of satellites, drones and sensors \cite{earthobs, earthag}.
Automated extraction of contextual information or insights from such geospatial data is important for many applications including autonomous driving, telecommunications, and disaster response. While for humans it is easy to recognize objects in images with a few examples, algorithms require massive amount of training data to carry out similar tasks. As an example, benefiting from large-scale labeled datasets like ImageNet \cite{imagenet}, 
image classification on photographic images using deep learning has advanced tremendously in the last decade \cite{dlreview} and eventually surpassed human performance on certain tasks \cite{resnet}. 
Similar machine learning (ML) techniques have been adapted to Earth science \cite{dlearth} for enhancing spatial and temporal resolution of imagery \cite{deepsum}, classifying hyperspectral images \cite{raczko2017}, and reconstructing missing data \cite{Zhang2018}. However, scalable adoption of ML approaches to solve real-world problems using geospatial data remains cumbersome due to the following challenges.

Firstly, unlike natural images which consist of Red (R), Blue (B), and Green (G) bands, geospatial data can be highly heterogeneous. Many satellites also record microwave, ultraviolet, infrared images, 
or even variations in gravity.
These multi-modal remote sensing signals often demand far more sophisticated processing compared with conventional photography. For instance, Synthetic Aperture Radar (SAR) \cite{sar}, which measures the back-scattered radar signal emitted from spacecraft in amplitude, phase, and polarization, requires specialized pre-processing steps to become usable by average users. Geospatial data also come with different spatial resolutions and timestamps. Thus it takes significant amount of effort to harmonize and align different datasets.

Secondly, existing ML models often cannot be easily adapted to multi-spectral geospatial data. By default, models created for image analysis tasks mostly assume the input to be in RGB format or even grayscale. However, geospatial data often come from multi-spectral data acquisition. Specific tasks may leverage vital information in channels beyond RGB. For example, near infrared (NIR) channel is sensitive to moisture content of plants and important for land cover classifications \cite{dlearth}. Thus, ML models need to be carefully designed to fit the multi-channel data format of geospatial data.

Thirdly, geospatial data are not efficiently accessible for ML applications. Satellite images, for example, are often hosted in a data repository in the form of individual files representing the data of a grid tile on Earth surface, typically with a size of ~1 gigabyte (GB) per file. One can search for data by an area of interest (AoI, a polygon) and data tiles which intersect with the AoI are returned. This design raises a problem: even if only one kilobyte (KB) of data from this AoI is needed, the user still needs to download/transfer the entire 1 GB tile and take additional steps to extract this one KB of data. This approach is extremely inefficient, especially for the deep learning methodology where small random data batches are fed to the models instead of a whole chunk. Towards this end, recent progresses include the creation of deep learning datasets based on high resolution satellite images \cite{deepsat, spacenet} for tasks like land use classification and building detection. However, if the application is different from what is intended for these datasets, or just to test the models  outside the coverage of these datasets, data accessibility remains a big obstacle.

Lastly, the time-to-value for geospatial data applications is still far from ideal given the above challenges. For example, Energy \& Utility companies are eager to use satellite observations to monitor the status of their assets and detect anomalies around them. In the case of vegetation management, having the right ML classifiers that can identify tree species and track how trees encroach power line assets can improve decisions on when and where to trim. However, it is a challenging task to relate a company's proprietary labels of assets and trees to the vast amount of satellite images, which are in completely different formats and repositories.
A similar example can be found in precision agriculture where ML models trained on remote sensing imagery can help with crop type identification and fertilization scheduling \cite{dlag}.

Indeed, while data labels of various assets exist in different industries, common challenges these industries face are how to assemble the training data that is usable for ML models, mitigate data quality issues, choose the right ML models, and integrate ML models into daily operations.

\begin{figure*}[t]
    \centering
    \includegraphics[width=0.95\textwidth]{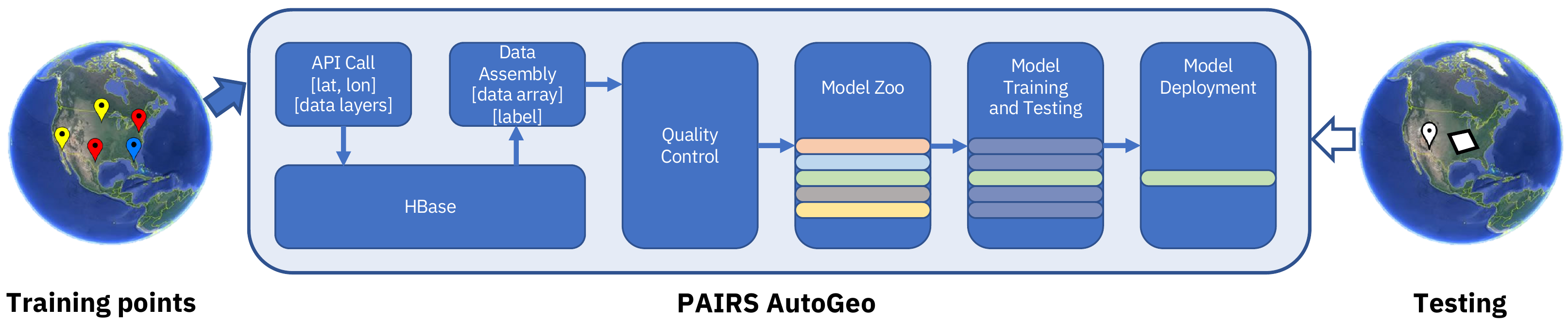}
    \caption{Process flow of the \algname framework for machine learning on geospatial data.}
    \label{fig:framework}
\end{figure*}

In this paper, we present an automated machine learning framework for massive geospatial data named \emph{\algnamens} (Fig.~\ref{fig:framework}) and illustrate how it may substantially lower the complexity and the barriers encountered by geospatial ML model development.
Currently, if a user wants to carry out an ML-based classification, it will have to search for the data from various repositories, download/transfer the data, spatially and temporally align different data layers, crop the data around the labeled location, train ML models and validate the model performance. Many of these tasks require not only data science and image processing skills but also detailed knowledge of the geospatial datasets, which restricts such endeavors to highly specialized individuals. With \algnamens, the complexity in data processing and model development is masked via automation, thus lowering common barriers to entry for geospatial ML industrial applications.
The only required input from the user is a text (JSON) file that contains the Global Position System (GPS) coordinates and  corresponding labels. Our contributions are summarized as follows:
\begin{itemize}
    \item We make geospatial data more easily and efficiently accessible to general public based on PAIRS geospatial platform \cite{pairs}, where users are relieved from heterogeneous pre-processing requirements and data can be queried for areas of interest only.
    \item We automate the data preparation for model training to remove the burden of gathering geospatial data from various data sources and locations. The input from users is reduced to only a list of GPS coordinates and corresponding labels.
    \item We include data quality control on the assembled data.
    \item We build customizable models that can inherently work with heterogeneous geospatial data rather than being limited to conventional RGB format.
    \item We integrate the solution into an end-to-end framework to speed up the research on geospatial data and shorten the time-to-value.
    \item We create a simple framework that can easily be adopted to various industry solutions where geolocated labels are abundant but the lack of geospatial data and skills prevents value from being materialized.
\end{itemize}

The paper is organized in the following way. In Section~\ref{sec:related} we summarize prior efforts in addressing the challenges in machine learning on geospatial data. We discuss the characteristics of geospatial data and its labeling in Section~\ref{sec:geospatial}. In Section~\ref{sec:method} the framework is explained in detail. As a demonstration, we present the process of training and testing a tree species classification model in Section~\ref{sec:exp}.

\section{Related Work}\label{sec:related}

\subsection{Geospatial data platforms}

Conventional geospatial data are typically stored as files in object stores managed by either data producers like European Space Agency (ESA), NASA, and United States Geological Survey (USGS), or commercial data stores such as Amazon Web Services (AWS) \cite{aws}. Each file is maintained in its original projection, spatial resolution and data format. A data scientist is required to develop a pipeline to search, download and crop the tiles that contain the AoI. The data curation process (spatial and temporal alignment, harmonization, and aggregation) is left to the user who needs to load all datasets 
and run the analytics. 
While file-based platforms may be good for infrequent data exchange, a more efficient data organization and data retrieval strategy is necessary to support ML solutions for industrial applications.
Google Earth Engine (GEE) \cite{googleearth} reduces the overhead by storing the data in smaller tiles and enabling querying by regions, however, except for machine learning using data from individual pixels, training data still have to be transferred out of GEE for more complex analytics and modelling  \cite{ce}.

\subsection{Geospatial datasets}
Specialized geospatial datasets have been assembled to classify land cover \cite{deepsat, aid, nwpu} and urban areas \cite{fmov} or to recognize roads and buildings \cite{spacenet}. Each of these datasets targets specific applications and requires new training data to be curated from similar data sources. 
However, they cannot be conveniently adapted to other applications, nor can they be integrated with other data sources (e.g., tree species labels from utility companies).

\subsection{Machine learning on geospatial data}
Machine learning on geospatial data is a developing field\cite{dlearth}. Models traditionally rely on random forest classification or support vector machines \cite{raczko2017}, but more recently deep learning techniques have been applied.
Taking tree species classification as an example, models have been developed based on satellite images\cite{treesp1, treecnn, wildfire}, Light Detection And Ranging (LiDAR) data \cite{treesp2, treesp3}, and SAR data \cite{sarsp}. While the performance of these models can be tuned based on carefully selected labels from well-maintained forest plots and high quality images, how to scale up the learning to more accessible labels and data sources remains an open question.

\subsection{AutoML platforms}
Most of the cloud service companies (e.g., IBM, Amazon, Google, Microsoft) also offer automated machine learning (AutoML) services (for example, image classification), with which the user drops images together with labels and receives a trained model for classification. Today, these services are universally based on RGB images and normally require the user to provide image data for the training. Existing AutoML platforms are not optimized for geospatial data given its heterogeneity. Thus new solutions are in need.

\section{Geospatial Data}\label{sec:geospatial}

\subsection{Geospatial data at a glance}

Satellite images are good examples to illustrate the properties of geospatial data. Commonly, they are acquired by spaceborne sensors in a top-down view capturing, for example, the top of tree canopies. 
While satellite images have some similarities with regular photographic images based on the RGB color model, they have three distinctive characteristics:
\begin{enumerate}
    \item \emph{Multi-spectral} data, with information acquired in microwave, ultraviolet, visible and/or infrared bands. 
    \item \emph{Localized} data, where each pixel of an image is associated with a certain location on Earth surface at a certain moment of time. Images from different satellites or acquired at different times can be stacked on top of each other and used together for ML. In contrast, stacking regular RGB images is usually impossible.
    \item \emph{Massive} data, with many terabytes (TB) of data generated every day. For example, the Sentinel program of ESA alone produces 18 TB of data daily \cite{esa}.
\end{enumerate}

The multi-spectral acquisition can enable detection of vegetation, clouds, and thermal sources \cite{esa}. Some of the hyperspectral satellites can offer hundreds of spectral bands \cite{treesp2}, providing extremely rich information of the observed area. 
Satellite images are often taken periodically over the same location. Upon proper registration to geolocations, they allow change detection in vegetation and ecological processes. However, it is crucial to have geospatial machine learning services that can process the data properly, deal with the multi-spectral and heterogeneous properties, and efficiently scale up to massive data sizes.

\subsection{Geospatial data labeling}

Manual labeling of large volume of data, such as ImageNet \cite{imagenet} or COCO \cite{coco}, has led to impressive improvements in image classification and object detection. 
By the nature of geospatial data, besides the quality of labels, the localization accuracy of labeled objects is also important. The localization accuracy is determined by the GPS localization accuracy and by the technical expertise of the person acquiring such labels. 

Labeling of geospatial data can be available at vastly different scales and quality. For proof of concept studies, researchers often manually label a small set of data to support model development \cite{Hogland2018}. Utility companies have tree species and vegetation status labels across their local service territory. Labeled datasets have been created to drive the development of ML models like SpaceNet \cite{spacenet} and Functional Map of the World \cite{fmov} at selected locations.

Government agencies have created labeled data such as land cover classifications, for example, Cropscape \cite{cropscape} in United States and Corine data sets \cite{corine} in Europe. Many of these datasets support agricultural research or tracking changes in land use (e.g., water, bare land, urban area). These classifications have already found applications in identifying construction sites, localizing and tracking deforestation, and monitoring human activities in remote areas.

Crowd-sourced labeling in efforts like OpenStreet Map \cite{osm}, where buildings, streets and urban areas are labeled by volunteers, can cover a large portion of the globe and is widely used. One caveat is that the quality of crowd-sourced labels can vary depending on the experience of the volunteers, the quality control process, and update frequency \cite{conradchange}.

Although labels for geospatial data are abundantly available, there often exists a gap between the labeled data from various sources and the corresponding geospatial data to overlap with. A solution that can effectively integrate these labels with massive geospatial data and speed up insight discovery is discussed next.

\section{Automated Machine Learning for Geospatial Data}\label{sec:method}

\begin{figure}
    \centering
    \includegraphics[width=0.87\columnwidth]{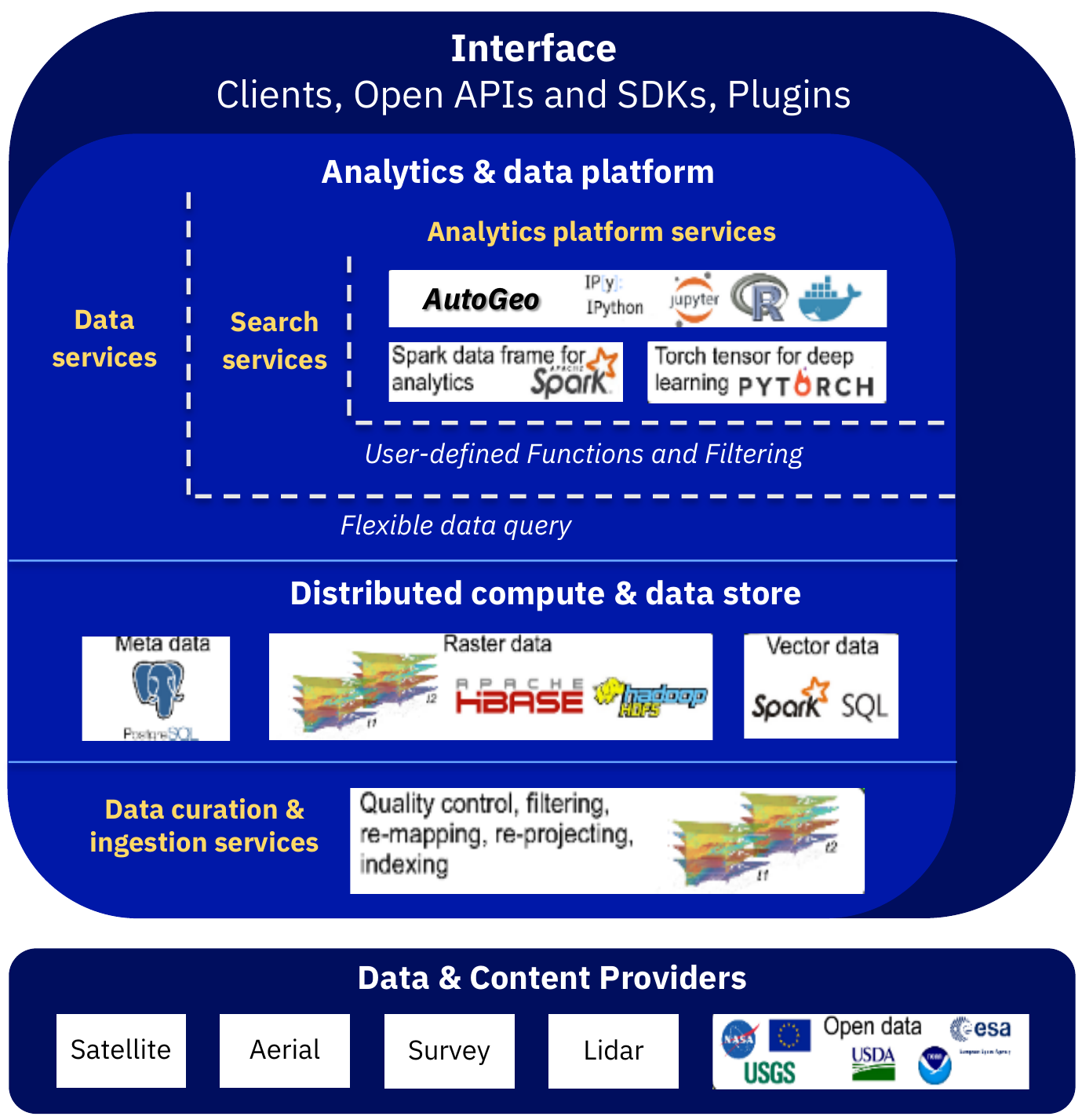}
    \caption{System architecture of PAIRS Geoscope platform including \algname as a module of the analytics platform services.}
    \label{fig:pairs}
\end{figure}

Here a framework called \emph{\algnamens} is designed to automate the workflow of machine learning on heterogeneous geospatial data and to minimize the workload on end users. 
The overall architecture is illustrated in Fig.~\ref{fig:framework}.
The approach largely reduces the user input to a simple JSON that contains labeled GPS coordinates. The generation of training data by gathering data around the coordinates is automated by leveraging PAIRS geospatial platform that serves as both a data repository and an in-data computational environment for ML models. 
\algname eliminates the requirements on users to create training data, setup and train ML models. Without loss of generality, we illustrate the workflow of \algname below with a classification task.

\subsection{Automated data preprocessing}

As discussed in Section~\ref{sec:geospatial}, geospatial data require special consideration in curation and preprocessing. While traditional file-based data platforms leave the burden of data collection to the end users, automated data preprocessing during data ingestion and efficient ad-hoc data retrieval are implemented on \emph{PAIRS Geoscope} \cite{pairs, pairs2, pairs3}. 
PAIRS stores geospatial data on a common map projection and a set of nested common grids. Fig.~\ref{fig:pairs} depicts the overall architecture of PAIRS Geoscope. The data curation engine preprocesses all data sources before  uploading so that all the data layers are automatically aligned both spatially and temporally, freeing the user from manually registering different data sources, which can become intractable.

\begin{lstlisting}[language=json, firstnumber=1, label={lst:query}]
query =	{
    "layers" : [{"type" : "raster", "id" : "50060"}, 
                {"type" : "raster", "id" : "50061"}, 
                {"type" : "raster", "id" : "50062"}, 
                {"type" : "raster", "id" : "50063"}
                ],
    "spatial" : {"type" : "point",
                 "coordinates" : [lat_1, lon_1]
                },
    "temporal" : {"intervals" : [{"snapshot" : "2018-01-29T12:00:00Z"}]
                 }
    }
\end{lstlisting}

Data can be queried for a specific region through the ``Data services'' API call \cite{pairsquery}, and only the data of interest 
will be returned. This has a unique advantage over traditional file-based geospatial data systems since one can query only 1 KB of data if needed without being forced to download a 1 GB tile and then crop out 1 KB of information. As shown above, to query data from PAIRS \cite{pairsweb}, the user only needs to assemble a JSON string which specifies: 1) data $\mathrm{layers}$ to query from, 2) the $\mathrm{spatial}$ span of interest (a point or an area), and 3) the $\mathrm{temporal}$ range. Once the query is submitted to the server, requested data will be retrieved from the big data store. 

\subsection{User input}
For a typical customized classification task, one has to gather data of interest for both the images and the associated labels. Specifically for geospatial data, a user often acquires images from massive data sources provided by government agents such as USGS and ESA. As an example, National Agriculture Imagery Program (NAIP) \cite{NAIP}, which is acquired in four optical bands (RGB-NIR) at a spatial resolution of $0.6\sim1$ m, 
is provided as tiles of 50 km lateral coverage. 
To generate a training label/feature dataset for a large number of locations across contiguous United States, it will require a user to sift through $\sim$60 TB of data. With an exceptional speed of 0.5 GB/s, this would take one and a half days to download all the data. It will take another week if conventional hard disk drive are used at 0.1 GB/s I/O speed to search, load into memory, and generate the training dataset. 
With \algnamens, the user's data preparation effort is reduced to one submission of a list of GPS locations (latitude and longitude) and corresponding labels, significantly reducing the data needed to be handled from TB to KB. As shown in the sample query below, the list is provided to \algname to construct API calls and assemble corresponding data.

\begin{lstlisting}[language=json, firstnumber=1]
query =	{
    "layers" : [...],
    "spatial" : {"type" : "point",
                 "coordinates" : [lat_1, lon_1;...;lat_N,lon_N]
                },
    "temporal" : {...}, 
    "model" : {"label" : [y_1;...;y_N],
               "window_size" : 32,
               "filters" : {"ndvi" : {"min" : 0.0}
                           }
              }
    }
\end{lstlisting}

\subsection{Data assembly}
For each of the GPS locations shown in the query above, a $k \times k$ pixel bounding box centered at the specified location is defined, and pixel values within the bounding box are parallelly queried from PAIRS' massive data store. The user can specify the window size $k$, which data layers are needed ($c$ layers) and when the labels are created. 
\algname will extract the data for each labeled point by matching the timestamp of the labels with corresponding acquisition time of the data layers. Users may define a search time window as part of the \texttt{"}$\mathrm{temporal}$\texttt{"} argument in the query. If a match cannot be found, \algname automatically searches the closest timestamp within the search window for which a data layer is available. The data retrieved is a $k \times k \times c$ array for each GPS location, and repeated for the whole list.

\subsection{Data quality control}
The quality of the training data is essential for the performance of ML models. For geospatial data, the data quality can be affected by: 1) errors in geo-location (GPS location is misplaced at wrong coordinates), 2) mismatch in the timestamps between when the labels are generated and when the data are acquired (for example, a newly constructed building which was labeled as ``bare land'' in an earlier time), and 3) erroneous data because of sensor issues or weather conditions. To filter out the invalid data, statistical measures ($\min$ / $\max$ / $\mathrm{mean}$ / $\mathrm{std}$) of the data or their derived composite indices can be used. Such functions can also be integrated in the API calls applying User Defined Functions (UDFs) \cite{pairsweb} as exemplified in Section~\ref{sec:exp}.

\subsection{Model training and testing}
Once the training data are generated and quality checked, machine learning models can be trained and tested. A set of predefined models suitable for geospatial data are provided in the ``model zoo''\footnote{Potentially more advanced Network Architecture Search (NAS) services can be built into \algname as well, so that the network can be further customized to the dataset.}, including random forest models \cite{randomforest} and deep learning models like (modified) ResNet \cite{resnet}. The user can choose a preferred model provided by the model zoo, or request to train a set of models and choose the best performer. A subset of the training data is held out to test the models and evaluate the performances. The hands-free model training and testing process makes ML models more accessible to general users who are not ML specialists.

\subsection{On-demand ML service}
After a ML model is trained, the user can make queries to apply the model at new locations by simply submitting a request with a GPS location (single point) or a bounding box (area) of interest. \algname goes through the same data processing steps to get the input data from PAIRS and feeds the data to the model. For the case of a bounding box, the area will be split into tiles of size $k \times k$, and the trained model will be applied to each tile. The model output after being assembled is returned to the user as the query result.

In this way, \algname can be thought of as an ``on-demand'' ML service  as it automates many manual steps which otherwise need to be carried out by the end user for ML on geospatial data: data downloading from various central repositories, common projection and spatial alignment, cropping the data using bounding boxes with the label locations being the center points, training and running the ML models, and validating the performance of ML models. With \algnamens, one can more easily train and deploy models on larger and more diversified datasets on a global scale, which may lead to improved generalization of ML models.

\section{Experiment}\label{sec:exp}

\begin{figure*}[t]
    \centering
    \includegraphics[trim={0 1cm 0 1cm},clip,width=0.8\textwidth]{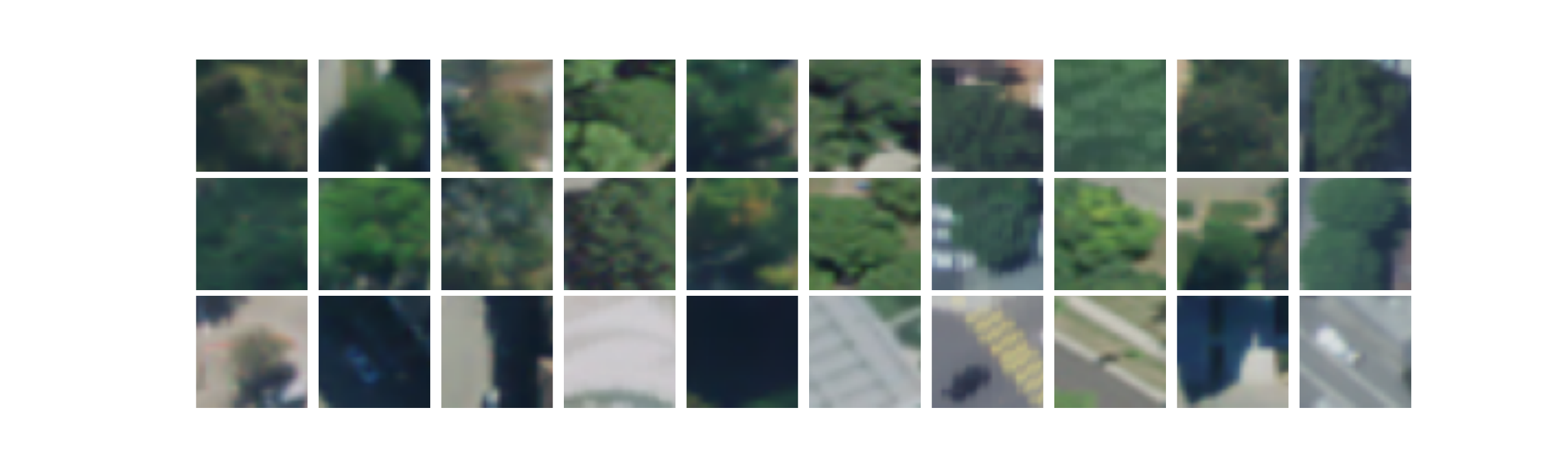}
    \caption{Sample images of different tree species. From left to right, each column represents one tree species in Table~\ref{tab:tree_data} in alphabetical order. The first two rows show good samples of trees, while the last row depicts bad samples of trees. Each image is $32 \times 32$ pixels with a spatial resolution of 0.5 m.}
    \label{fig:tree_samples}
\end{figure*}

To demonstrate the workflow of \algnamens, we consider a use case where tree species are identified using aerial imagery. Tree species classification is important for utility companies as they need to prioritize trimming schedules based on the growth rates of different tree species for vegetation management.
Urban trees can regulate the local environment by providing green zones and lowering the ambient temperature. Tree shades can lower cooling cost during summer time \cite{urbantree} and have a positive impact in minimizing pollution\cite{urbanpollution}. 
Given the ecological and environmental impact that trees have on local communities, volunteer groups are maintaining openly accessible tree inventories in cities 
\cite{urbantreenyc, opentree}, which can be used for training tree classification models.

\subsection{Dataset} 
We use open-source tree species labels acquired by municipalities and states \cite{opentree}. The data contain information about tree species, tree locations, and various other tree ecological information (e.g., crown size, health). In order to cover various tree species, data have been assembled from four different states across United States: Texas (Austin, Dallas, and Houston), Michigan (Denver, Ann Arbor), California (San Francisco and San Jose, Berkeley) and Massachusetts (Boston). The dominant tree species for each region are selected based on the occurrence frequency, and the final dataset consists of 10 tree species as listed in Table~\ref{tab:tree_data}. 
This list of trees with their GPS locations is the only ``training'' data needed from a user.

NAIP imagery with four bands (RGB-NIR) is used as the input data for tree species classification. The list of tree locations is provided to \algnamens, and for each location a $32 \times 32 \times 4$ data matrix (resampled to a spatial resolution of 0.5 m in PAIRS) is queried out from the PAIRS data store, with box size $k = 32$ pixels and number of layers $c = 4$. Fig.~\ref{fig:tree_samples} shows samples of both ``good'' and ``bad'' images of RGB bands for each of the tree species. Unlike standard image classification tasks where the samples are distinctly different (e.g. cats v.s. dogs), different tree species are very much alike, indicating the difficulty in tree species classification. There are no tree in the ``bad'' samples shown in the last row in Fig.~\ref{fig:tree_samples}, and they are probably due to errors in GPS locations or removal of trees after labeling, as discussed in Section~\ref{sec:method}. Mixing these ``bad'' images in the dataset would degrade the training data quality. 
In this case, Normalized Difference Vegetation Index (NDVI) \cite{ndvi},
\begin{equation}
    \mathrm{NDVI} = \frac{\mathrm{NIR} - \mathrm{Red}}{\mathrm{NIR} + \mathrm{Red}},
    \label{eq:ndvi}
\end{equation}
which is typically positive for areas covered by vegetation and negative for bare land \cite{kleindim}, can be useful to filter out the ``bad'' samples. Using such rule-of-thumb, we filter out all the samples with $\overline{\mathrm{NDVI}} < 0.0$ to improve the data quality. The final number of data points are listed in Table~\ref{tab:tree_data}. The labels are assigned in alphabetical order. For each of the ten tree species,  500 samples are randomly held out for validation and another 500 samples for testing, and the rest are used for training. Thus there are 118,750 / 5,000 / 5,000 samples in the training / validation / testing subsets, respectively.

\begin{table}[tb]
    \centering
    \caption{Dataset of tree species.}
    \label{tab:tree_data}
    \begin{tabular}{l c r}
         \hline
         Tree type & Label & \# points \\
         \hline
		Cedar\_elm           & 0 &  8,735 \\
		Crape\_myrtle         & 1 & 32,340 \\
		Hackberry            & 2 & 14,075 \\
		Honeylocust          & 3 &  4,583 \\
		Live\_oak             & 4 & 36,032 \\
		Maple                & 5 & 13,368 \\
		Metrosideros\_excelsa & 6 &  2,228 \\
		Oak                  & 7 &  4,667 \\
		Pecan                & 8 &  8,669 \\
		Sycamore             & 9 &  4,053 \\
         \hline
         Total              & & 128,750 \\
         \hline
    \end{tabular}
\end{table}

\subsection{Models}

Two baseline models are investigated: 1) the random forest (RF) classifier and 2) the deep learning model ResNet, which are part of the model zoo. The two models illustrate two ML training  approaches: 1) hand-generated features (RF) and 2) end-to-end data-driven deep learning (ResNet).

\subsubsection{Random forest models}


Random forest classifiers take a list of features as input and determine the classification from an ensemble of decision trees (estimators). For this kind of model training, hand-generated features need to be defined before feeding the data to the model. Within the query, the user can specify the features required to be computed from the raw data, and proper functions are called internally to convert the raw data into a feature vector for each of the GPS locations. For the use case of tree species classification, texture features including mean, standard deviation, and horizontal gray level co-occurrence matrix (GLCM) contrast are computed on Red, Blue, Green, and NIR bands (indicated by layers 0, 1, 2, 3 in the example below) \cite{Hogland2018}. In addition, mean and standard deviation of NDVI are also added as features. 
The code below specifies the features requested when the query is submitted.

\begin{lstlisting}[language=json, firstnumber=1, captionpos=b, label={lst:rf}]
query =	{
    "layers" : [...],
    "spatial" : {...},
    "temporal" : {...}, 
    "model" : {"mode" : "train",
               "id" : "tree_species_random_forest",
               "architecture" : "random_forest",
               "label" : [y_1;...;y_N],
               "window_size" : 32,
               "filters" : {"ndvi" : {"min" : 0.0}
                           },
               "params" : {"features" : {"mean" : [0,1,2,3,"ndvi"],
                                         "std" : [0,1,2,3,"ndvi"],
                                         "glcm:contrast" : [0,1,2,3]
                                         }
                          }
              }
    }
\end{lstlisting}

\begin{figure}
    \centering
    \includegraphics[trim={0 1mm 0 4mm},clip,width=0.85\columnwidth]{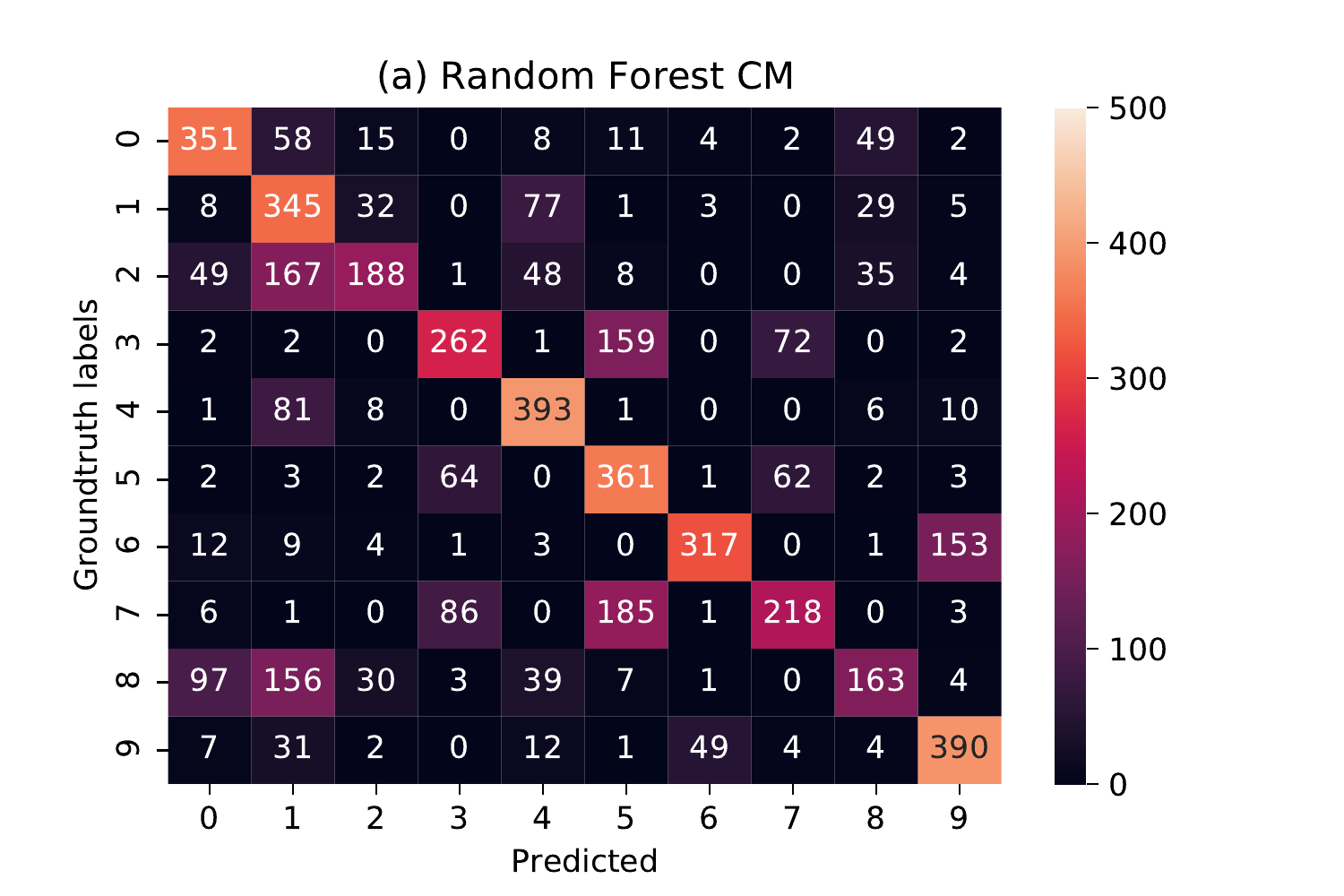}
    \includegraphics[trim={0 1mm 0 1mm},clip,width=0.85\columnwidth]{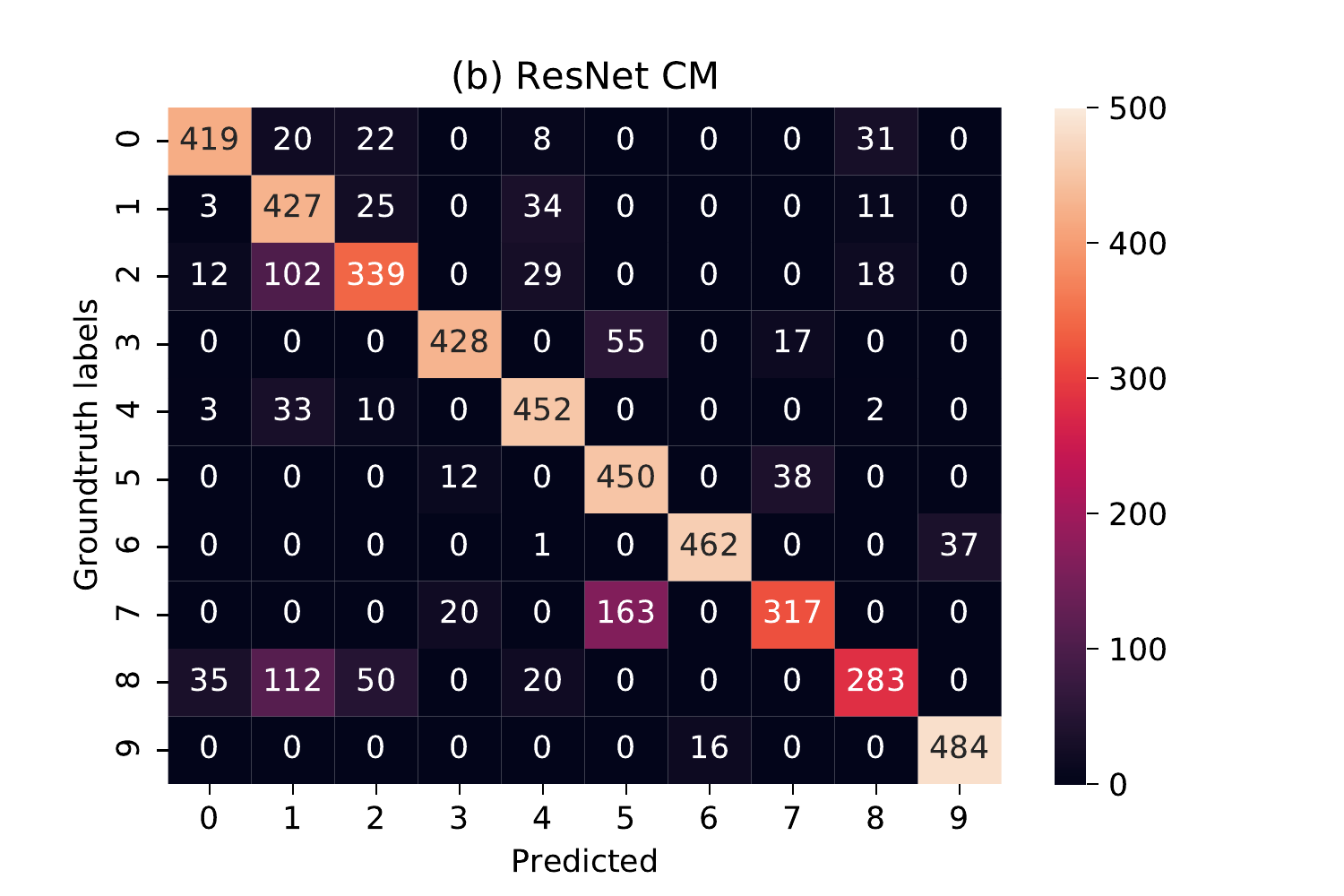}
    \caption{Confusion matrix (CM) of the classification results produced by (a) random forest classifier and (b) ResNet classifier. Each class has 500 testing samples. The overall classification accuracy for the random forest classifier is $59.8\%$, while ResNet achieves substantially higher accuracy of $81.4\%$.}
    \label{fig:cm}
\end{figure}

After the input features are prepared, a random forest classifier is trained on the training split. By default, \algname conducts grid search for $\mathrm{n\_estimators}$ and $\mathrm{max\_depth}$ (other parameters can be searched too upon request) based on the model performance on held-out validation set. The best model is then returned and tested on the test set. The final accuracy on the test dataset for the random forest classifier is $59.8\%$ for $\mathrm{n\_estimators} = 200$ and $\mathrm{max\_depth} = 15$. The confusion matrix in Fig.~\ref{fig:cm}(a) shows obvious mixing of classification between some tree species, suggesting that the representation capability of hand-generated features is limited.

\subsubsection{Deep learning models}

Deep neural networks have been dominating the task of classifying images with the help of large labeled datasets like CIFAR \cite{cifar} and ImageNet \cite{imagenet}. The neural networks are trained end-to-end without the need of hand-generated features like RF models. Instead, the features which are important for recognizing image categories are learned directly from the data. However, the majority of the existing image classification tasks are limited to RGB bands, and most predefined networks in standard frameworks are designed to take only those three bands as input.

In \algnamens, we customize the networks by allowing \emph{flexible} input formats so that the models can be applied to arbitrary number of data layers as requested by a user. This design is crucial as it enables the users to query many data layers from PAIRS and build customized deep learning models on demand.
As an example, here we apply a modified \emph{ResNet34} \cite{resnet} that can take inputs with 4 bands so that all the 4 channels of NAIP imagery can be used. 
The training can be initiated by the query below.

\begin{lstlisting}[language=json, firstnumber=1, label={lst:train}]
query =	{
    "layers" : [...],
    "spatial" : {...},
    "temporal" : {...}, 
    "model" : {"mode" : "train",
               "id" : "tree_species_resnet",
               "architecture" : "resnet",
               "label" : [y_1;...;y_N],
               "window_size" : 32,
               "filters" : {"ndvi" : {"min" : 0.0}
                           },
               "params" : {...}
              }
    }
\end{lstlisting}

The default training configurations for ResNet34 are as follows. An SGD optimizer with a mini-batch size of 512 is used, with a momentum of 0.9 and a weight decaying of 0.0005. The learning rate starts from 0.1 and is divided by 10 every 100 epochs and in total the model is trained for 400 epochs. A CrossEntropy loss is used for optimization (class-weighted CrossEntropy loss or Focal loss \cite{focal} can also be used, but the results turn out inferior in this case).
We apply random horizontal flipping and random cropping for training, while for testing no data augmentation is used. The implementation is based on Pytorch.

The ResNet34 model achieves an accuracy of $81.4\%$, outperforming the random forest model by a large margin. The confusion matrix in Fig.~\ref{fig:cm}(b) also has less mixing of classification. This implies that the learned features from the data are better than the hand-generated features in the task of tree species classification.
The classification of ``Hackberry''(2), ``Oak''(7) and ``Pecan''(8) falls short in both models, which may result from errors in the labeling. More domain specific knowledge may help to improve the data quality control.

\subsection{Inference on demand}

\begin{figure}
    \centering
    \includegraphics[width=0.95\columnwidth]{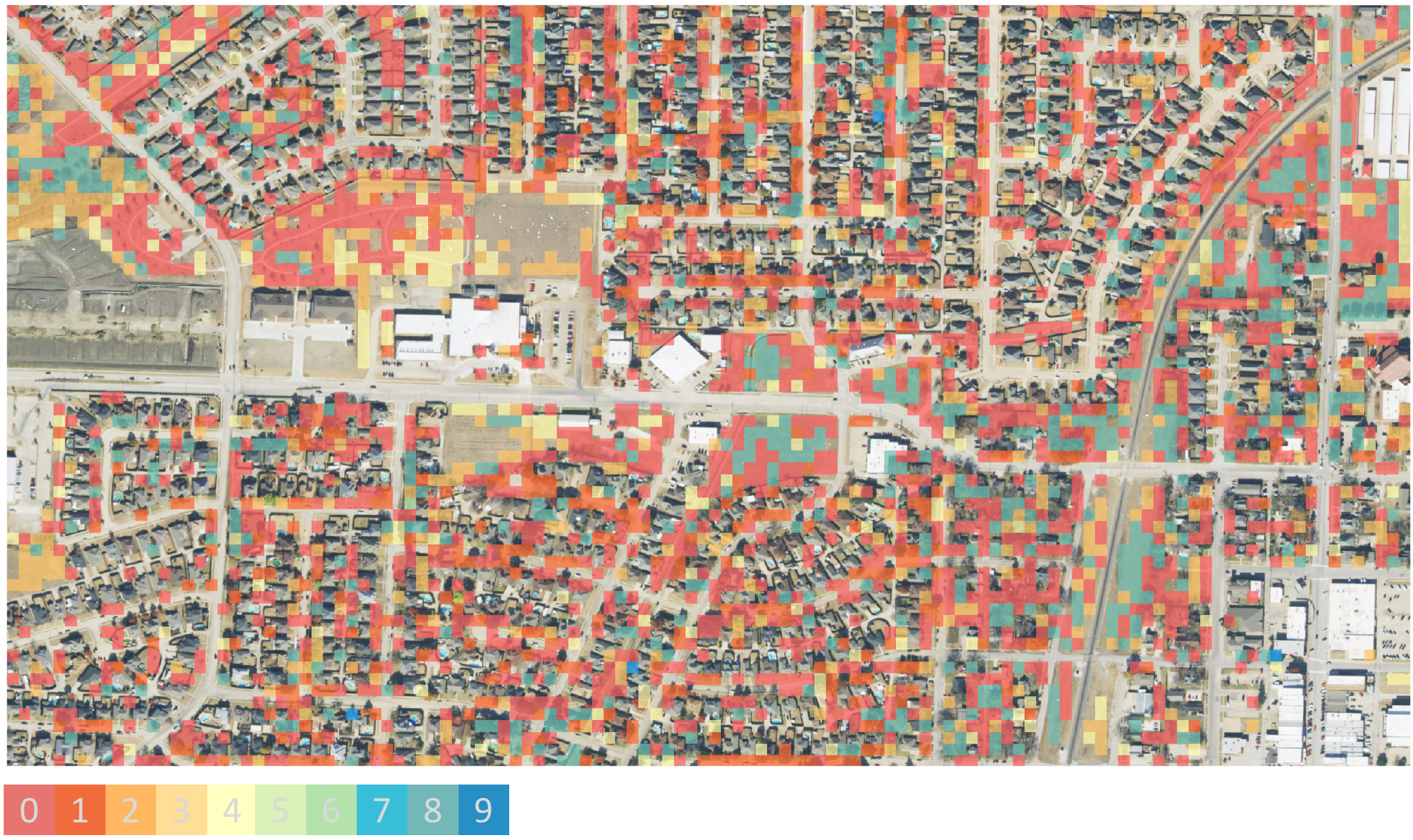}
    \caption{Classified tree species overlaid on a NAIP map. Each color corresponds to a tree species listed in Table~\ref{tab:tree_data}. Each block covers $32 \times 32$ pixels at a resolution of 0.5 m.}
    \label{fig:tx}
\end{figure}

Once the model has been trained with the list of GPS locations and labels provided by the user, further inference can be done on demand by requesting \algname to classify either a single GPS location (lat, lon) or an area defined by a pair of GPS points representing two diagonal corners of the bounding box. The following query illustrates a sample query JSON string to run inference with \algnamens. The query JSON keeps the data layers, data filters and model id the same as the training query, but the model mode is specified as ``test''. 

After the query is submitted, \algname will: 1) retrieve the required data from the PAIRS data store, 2) split the data into $k \times k$ sized tiles, 3) filter the data by the specified rules (and the classification results of these filtered tiles will be set to $\mathrm{None}$), 4) feed the rest of the tiles to the model and get classification results for each one of them, 5) assemble the individual results into an 2D array, and 6) return the final result to the user.

\begin{lstlisting}[language=json, firstnumber=1, label={lst:test}]
query =	{
    "layers" : [...],
    "spatial" : {"type" : "square",
                 "coordinates" : [lat_1, lon_1, lat_2, lon_2]
                },
    "temporal" : {"intervals" : [{"snapshot" : "2018-01-29T12:00:00Z"}]
                 }, 
    "model" : {"mode" : "test",
               "id" : "tree_species_resnet",
               "architecture" : "resnet",
               "window_size" : 32,
               "filters" : {"ndvi" : {"min" : 0.0}
                           }
              }
    }
\end{lstlisting}

Fig.~\ref{fig:tx} shows a  map where an area near Dallas, TX is queried for inference for tree species classification. Although there are no labels available to quantify the model performance, unsurprisingly, the majority of the species ``Cedar\_elm''(0), ``Crape\_myrtle''(1), ``Live\_oak''(4) and ``Pecan''(8) are in fact among the dominant species in Texas.

\section{Conclusion}
We introduced an automated machine learning framework specially designed for geospatial data called \algnamens. Compared to traditional approaches to develop machine learning models on geospatial data, \algname automates many steps of the process, including data curation, quality control and model training and testing. The framework requires a user to provide a simple text (JSON) file representing GPS locations and corresponding labels, and eliminates the workload on the user to collect training data and run the models. We demonstrate the framework with a use case of tree species classification using NAIP imagery as a data source, and achieve an accuracy of $59.8\%$ using a random forest classifier with hand-generated features and $81.4\%$ using a modified deep learning model ResNet. Such an automated framework can accelerate the adoption of machine learning techniques in different industries (Insurance, Banking, Agriculture, Forestry, Land Management, Oil \& Gas and Energy \& Utility) as it demands little technical knowledge of the intricate processing that is required to develop machine learning solutions on geospatial data today.


\clearpage
{
\small
\bibliographystyle{IEEEtran}
\bibliography{ref}
}

\end{document}